*Silvia Alma Piazzolla*, *Beatrice Savoldi**, *Luisa Bentivogli***

# *Good*, but not always *Fair*: An Evaluation of Gender Bias for three commercial Machine Translation Systems.

## Abstract

Machine Translation (MT) continues to make significant strides in quality and is increasingly adopted on a larger scale. Consequently, analyses have been redirected to more nuanced aspects, intricate phenomena, as well as potential risks that may arise from the widespread use of MT tools. Along this line, this paper offers a meticulous assessment of three commercial MT systems - Google Translate, DeepL, and ModernMT - with a specific focus on gender translation and bias. For three language pairs (English→Spanish, English→Italian, and English→French), we scrutinize the behavior of such systems at several levels of granularity and on a variety of naturally occurring gender phenomena in translation. Our study takes stock of the current state of online MT tools, by revealing significant discrepancies in the gender translation of the three systems, with each system displaying varying degrees of bias despite their overall translation quality.

## 1. Introduction

In the age of globalization and digital communication, Machine Translation (MT) represents one of the most largely employed language applications. In 2016, Google Translate was estimated to generate more than 100 billion words per day (Pichai, 2016), and its use has further increased ever since (Pitman, 2021). MT is integrated into e-commerce platforms such as eBay, search engines, and Bahdanau online chats, whereas online MT tools are deployed for a range of use cases by millions of people daily, e.g., for leisure, traveling, or work-related activities (Vieira et al., 2022).

Besides converging with the multilingual demands of our society, MT's exponential popularity can be largely attributed to the advent of neural approaches, which have advanced the state-of-the-art by achieving unprecedented performance (Sutskever et al., 2014; Chorowski et al., 2015; Vaswani et al., 2017). Such neural models are trained on vast amounts of parallel data,[1] from which they infer and generalize cross-lingual patterns and associations to perform a translation. Their strength lies primarily in such generalization capabilities, which has led to higher levels of fluency and accuracy for several types of linguistic phenomena (Bentivogli et al., 2016; Toral & Sánchez-Cartagena, 2017). This enhanced quality has enabled MT to meet the growing demand for seamless cross-lingual communication and to overcome language barriers in multiple sectors and settings, making it a popular technology.

Despite notable advancements in the field, however, MT still presents some critical challenges in cross-lingual transfer. Additionally, as MT technologies are deployed at scale and in a variety of contexts of use, their social and ethical implications have become the object of increasing discussions (Bender & Friedman, 2018). One such issue that has received growing attention in recent years – combining linguistic, societal, and ethical dimensions – is gender bias (Ullman, 2022; Lardelli & Gromman, 2022). In fact, MT models have been found to perform incorrect gender attribution,

---

[1] Source text aligned with its corresponding target translation.

| | | |
|---|---|---|
| * Silvia Alma Piazzolla | ** Beatrice Savoldi | ***Luisa Bentivogli |
| Department of Humanities | Machine Translation | Machine Translation |
| University of Trento | Research Unit | Research Unit |
| silviapiazz@gmail.com | Fondazione Bruno Kessler. | Fondazione Bruno Kessler. |
| | bsavoldi@fbk.eu | bentivo@fbk.eu |

reproduce stereotypes, and overuse masculine forms in translation (Savoldi et al., 2021). Besides the scientific community, such an issue has gained relevance also for the general public, and key MT providers have taken several steps to mitigate bias and provide fairer translation outputs (Kuczmarski, 2018; Naik et al., 2023). However, despite their potential societal impact and massive deployment, few studies have been dedicated to auditing such commercial MT systems (Monti, 2017; Prates et al., 2018; Rescigno et al., 2020).

Accordingly, this paper contributes to this line of inquiry by taking stock of the behaviors of three widely employed commercial MT systems regarding gender bias: Google Translate, DeepL, and ModernMT. To this aim, we focus on the translation between English-Italian, English-French, and English-Spanish – as described in §2 – as representatives of the main cross-lingual complexities that can lead to the emergence of biased output. On this basis, we perform a focused evaluation at several levels of granularity on natural, gender translation phenomena, thus assessing how and to what extent the three MT models tend to be biased and underrepresent feminine forms. In particular, in our analysis we consistently find that i) overall translation quality does not directly relate to gender translation ability; ii) these models show acceptable gender translation ability when gender translation is not ambiguous, whereas ambiguous cases lead to a strong masculine skew; iii) occupational nouns are among the words mostly affected by gender bias in translation; finally, iv) while all three investigated systems present a masculine bias, among them DeepL emerges as the model that marginally better handles feminine translation.

## 2. Gender (bias) in translation and MT

To delve into the issue of gender bias, we first discuss relevant differences in how gender is expressed across languages (§2.1), and then outline related, existing work on the topic in MT (§2.2).

### 2.1 Gender across languages

Different languages express references to the social, extra-linguistic reality of gender in different ways. When it comes to human referents, languages vary in terms of how they mark gender and the linguistic strategies they employ to do so (Stahlberg et al., 2007). Some languages, like many Indo-European ones, are known as "grammatical gender" languages. In these languages, gender is a formal and semantic distinction that applies to various parts of speech (Konishi, 1993; Corbett, 2013), which can be inflected by means of feminine or masculine morphology (e.g it: la/il nuova/o amica/o - *the new friend*). On the other hand, other languages such as English are described as "notional" or "natural gender" languages (Curzan, 2003), whereby gender is expressed mostly by pronouns (e.g.: he/she is a friend), or differentiated at the lexical level (e.g., mum, dad).[2]

These general distinctions highlight that gender expressions in languages can differ, leading to mismatches and gender-specific associations. These differences become particularly relevant when translating from notional gender languages to grammatical gender ones. Since the former group of languages avoids gender inflection, human translators may inadvertently introduce gender biases in their translations. They might choose to use a masculine gender form instead of a feminine one, or vice versa, influenced by implicit stereotypes or prejudices related to their language's means of gender expression (Nissen, 2002). For instance, when translating professional titles such as "doctor" or "nurse", there is a tendency to stereotype which roles are appropriate for men and women (Hamilton, 1988; Gygax et al., 2008; Kreiner et al., 2008).

These types of biases, however, are not limited to human translators and their translations. Automatic translation systems are also susceptible to reflecting gender disparities present in their training data (Savoldi et al., 2021, p. 849). Consequently, these biases can manifest in the outputs of MT systems, thus perpetuating and reinforcing societal biases.

---

[2] In our study, we focus on traditional feminine/masculine linguistic forms also due to the lack of resources representing emerging non-binary linguistic strategies. Although we discuss bimodal linguistic forms, we stress that social gender and gender identities are multifaceted and encompass multiple biosocial elements not to be confused with sex.

## 2.2 Related studies on gender bias in MT

With substantial advances in the quality of MT output, the research focus has shifted to other fine-grained aspects, complex phenomena, and areas of risks engendered by the deployment of MT tools. Among them is gender bias.[3] Indeed, MT systems encounter the same challenges faced in human translations when rendering references to the extra-linguistic reality of gender across languages that express it differently. In fact, whereas the English expression "*a dancer*" does not convey any gender differences, in a grammatical gender language such as Italian two alternatives are possible: the masculine "*il ballerino*" and the feminine "*la ballerina*".

Figure 1: Example of biased gender translation with Google Translate: 'Sexy dancer' → ballerina (F) / 'Top dancer' → ballerino (M). In the sentence, sexy dancer is translated as feminine regardless of the source masculine cue (boy).[4]

Accordingly, systems are faced with a so-called mapping of one-to-many forms in translations. When predicting the target gender for such ambiguous cases, current systems are skewed toward the generation of masculine forms by default (Schiebinger, 2014; Vanmassenhove et al., 2018; Monti, 2020), or translate gender in a stereotypical manner, such as rendering *nurse* and *secretary* as feminine, but doctors and the more prestigious *secretary of state* as masculine (Nissim & Pannitto, 2022, p. 117). Stereotypical translations can also be triggered by co-occurring adjectives, such as in association with attractiveness judgments, which can lead systems to override gender cues available in the sentence and pick the wrong gender form, as shown in Figure 1.

As attested by growing studies on the topic (Savoldi et al., 2021), problems related to gender translation in MT are the convergence of several factors besides cross-lingual differences and are also due to the fact that the data used to train the systems already contain gender imbalances, which results in systems learning and reproducing them. Overall, however, these biased tendencies in MT do not only constitute a technical problem, inasmuch as gender-related errors can deteriorate the performance of a system. Rather, they can under- and misrepresent socio-demographic groups, by also amplifying controversial gendered associations (Hovy & Spruit, 2016; Crawford, 2017).

On the one hand, to tackle the issue several bias mitigation strategies have been proposed (e.g., Saunders & Byrne, 2020; Stafanovičs et al., 2020; Choubey et al., 2021; Sharma et al., 2022), which may entail technical interventions on architectural choices and systems training data (Gaido et al., 2020; Escolano et al., 2021). On the other hand, substantial effort is put into auditing existing MT models, to shed light on the main weaknesses and help unveil the extent of biased phenomena. Toward this goal, dedicated benchmarks have been created to support focused analyses on how systems behave with respect to gender translation (Stanovsky et al., 2019; Bentivogli et al., 2020; Levy et al., 2021; Renduchintala & Williams, 2022; Currey et al., 2022; Rarrick et al., 2023).

---

[3] Note that gender bias is not a pathology exclusive to MT technology but has rather been attested across different aspects and conceptualizations – in a variety of Natural Language Processing tools; see Sun et al. (2019).
[4] The query dates to May 5[th], 2023.

While auditing MT represents a crucial step towards informing both mitigating strategies as well as users interacting with these systems, to the best of our knowledge only a few studies have so far inspected how commercial systems are affected by the problem. Namely, the pivotal work by Prates et al., (2018) detected how Google Translate, when confronted with the translations of occupational nouns, was not only skewed towards masculine forms, but also underestimated feminine frequency in such professions at a higher rate than actual US occupation data alone would suggest. Analogously, Monti (2017), and Rescigno et al., (2020) focused on Microsoft Bing, Google Translate, and DeepL with respect to the translation of sentences featuring occupational nouns and/or potentially gendered connoted adjectives (e.g., beautiful, famous). All in all, while crucial, all these studies have investigated gender bias with respect to a restricted set of lexical items by means of synthetic benchmarks, which consist of sentences with similar structures and feature limited linguistic variability (Savoldi et al., 2021). Instead, in this work we intend to contribute to this line of research by auditing commercial systems on naturally occurring and highly variable gendered phenomena. In this way, we can better grasp the systems' capabilities on real-world-like instances of gender translations and take stock of the current situation of commercial MT, whose capabilities incessantly evolve in the face of rapid technological advancements.

## 3. Experimental Settings and Methodology

Hereby, we describe the experimental settings of our analysis of gender bias in MT. Accordingly, we first present the three commercial MT systems audited in this work (§3.1). Then, we introduce the MuST-SHE benchmark employed in our experiments (Bentivogli et al., 2020; Savoldi et al., 2022) and its associated evaluation method (§3.2).

### 3.1. Machine Translation Systems

We compare the performance and ability to translate gender of three state-of-the-art neural MT systems: Google Translate,[5] DeepL,[6] and ModernMT.[7]

**Google Translate (GT)** is a web-based, free-to-user MT system developed by Google. First launched in 2006 as a Phrase-Based system (Och, 2006), it switched to a multilingual, neural-network-based architecture in 2016 (Wu et al., 2016; Johnson et al., 2017) achieving significant translation quality improvements on several major language pairs. As of today, it supports more than 130 languages and is possibly the largest existing MT provider.

**DeepL Translate (DL)** is a freely available translation service that was launched by the German company DeepL GmbH in 2017. It is one of the most widely used translation systems by more than half a billion people, which supports 31 languages. According to results reported in DL press releases,[8] as well as recent external evaluation studies (Tavosanis, 2020), the translation quality offered by DL might outperform other competitors (e.g., for en-it and en-de).

**ModernMT (MMT)** is an open-source translation technology developed via an EU-funded Horizon project and then launched into the market in 2018 (German et al., 2016). Including translation for 88 languages, MMT was created for translation production environments, both for fully automatic use as well as a back-end in interactive post-editing scenarios (Vashee, 2021a). As such - unlike GT and DL that are most widely employed by lay end-users - this model is particularly representative of the pipeline use of MT by translators.

---

[5] https://translate.google.com/
[6] www.deepl.com/translator
[7] www.modernmt.eu
[8] https://www.deepl.com/en/blog/20200206

Overall, such systems were chosen for our analysis in light of *i)* their popularity; *ii)* the inclusion of our three language pairs of interest (English → French/Italian/Spanish); *iii)* the representation of different scenarios of MT use*; and iv)* the different degrees of intervention made in terms of gender bias. In fact, concerning gender bias, it should be noted that since 2018 GT web interface provides two gendered translation alternatives for 'ambiguous' input queries (Kuczmarski, 2018), e.g., *cousin* → IT: *cugino/cugina*. Such a solution, however, only works for very short input consisting of a few words, and it is thus not available for longer sentences as those employed in our test set (see §3.2). However, to the best of our knowledge, MMT has not publicly set out any step towards making their systems more gender-inclusive and – unlike DL and GT (Monti, 2017; Prates et al., 2020; Rescigno et al., 2020) – has never been publicly assessed in terms of gender bias. Finally, DL online interface offers multiple translation suggestions for each input query, although these are not systematically intended to target gendered variants and rather more generally offer various synonyms, e.g., you are a bad friend → IT: sei *un pessimo amico*/sei *un cattivo amico*/ sei *una cattiva amica*).

### 3.2. Test Set and Evaluation Method

Automated procedures are typically employed to evaluate the quality of MT output, to scale the assessment of large pools of (input) source sentences that are automatically translated into the target language. Such an automatic evaluation procedure traditionally involves comparing the output of the MT model against collected human translations (i.e., *reference* translations) and computing their similarity using automatic metrics. Among them, the BiLingual Evaluation Understudy (BLEU) (Papinei et al., 2002) represents the *de-facto* standard MT metric,[9] which is based on the idea that the closer the MT output is to a human translation, the higher its quality. Thus, BLEU measures the overlap between words (or *n-grams*) appearing in both the reference translation and MT system output and frames their similarity with scores ranging between 0 and 1, typically converted into a percentage scale: the higher the score, the better the quality of the translation output.

Indeed, MT generic evaluation metrics such as BLEU serve as an indicator of translation quality as a whole, and variations in BLEU scores allow for comparison across systems to indicate better/worse overall performance (Callison-Burch et al., 2006).

|  | en-es | en-fr | en-it |
|---|---|---|---|
| Fem. | 551 (950) | 540 (898) | 515 (898) |
| Masc. | 555 (1027) | 529 (925) | 546 (1044) |
| Tot. | 1106 (1977) | 1074 (1823) | 1,061 (1,942) |

Table 1: MuST-SHE statistics. En-it, en-fr, and en-es number of segments split into feminine (fem.) and masculine (masc.) phenomena. In parentheses, the number of annotated gender-marked words per each gender form.

Such an aggregated global score, however, is not suitable for the precise evaluation of the systems' behavior for specific phenomena of interest, such as DL, GT, and MMT's ability to deal with gender and its translation. As previously discussed (§2.2), for fine-grained, focused automatic evaluation of gender bias, dedicated test sets are to be employed, and on which gender-sensitive ad-hoc metrics can be computed.

---

[9] See however Kocmi et al., (2021) for a review of other MT evaluation methods, including referenceless approaches.

**Test set.**

To evaluate our three MT systems, we employ the multilingual, gender-sensitive MuST-SHE benchmark (Bentivogli et al., 2020) and its annotated extension (Savoldi et al., 2022),10 both available for English→ French/Italian/Spanish. Built upon TED talks spoken language data, for each language pair MuST-SHE comprises ~1000 <source transcript, reference translation>11 segments aligned at the sentence level, featuring a variety of balanced masculine and feminine gender phenomena (see Table 1 for MuST-SHE segment- and word-level statistics). In the reference translations of the corpus, each target gender-marked word – corresponding to a neutral expression in the English source – is annotated with its alternative wrong gender form (e.g., en: the boy left → it: il<la> ragazzo è andato<andata> via.). As further discussed below (see: Evaluation Method), such a feature enables fine-grained analyses of gender realization. Because of its manageable size, MuST-SHE enables assessing gender bias across several informative dimensions annotated in the corpus:

- GENDER, which distinguishes translation performance for Feminine (F) and Masculine (M) forms, thus revealing a potential gender gap.
- CATEGORY, which differentiates between *i)* CAT1: first-person references that are translated in accordance with the speaker's linguistic expression of gender (e.g., en: *I am **a friend***, es: *soy **un amigo*** vs. *soy **una amiga***); and *ii)* CAT2: references translated in concordance with explicit gender cues in the sentence (e.g. en: *She is **a friend***, es: *es **una amiga***). These categories separate contextually ambiguous from unambiguous cases.
- CLASS & POS, which identifies if different gendered Parts-of-Speech (POS) are equally impacted by bias. POS can be grouped into open class (verb, noun, descriptive adjective) and closed class words (article, pronoun, limiting adjective).

Hence, we decided to rely on MuST-SHE as it represents the only multilingual, MT benchmark that allows to explore bias across numerous, and qualitatively different grammatical gender instances under authentic, natural conditions. Furthermore, the target languages covered in MuST-SHE (es, fr, it) are particularly suitable for proper comparisons, as they allow accounting for gender in languages with similar typological features. In our experiments, we consider all of the above-mentioned dimensions to explore gender translation at several levels of granularity. To do so, we follow MuST-SHE original evaluation method and metrics, whose robustness and reliability have been previously verified by Savoldi et al. (2022).

**Evaluation method.**

As mentioned above, in MuST-SHE reference translation each target gender-marked is annotated with its corresponding wrong gender-marked form. As put forth by Savoldi et al., (2022, p. 1809), such a feature enables pinpointed evaluations on gender realization in systems output by first computing *i)* **Term Coverage** (TC), i.e., the proportion of MuST-SHE annotated words that are generated by the MT system (regardless of their gender form), and on which gender realization is hence measurable,[12] e.g., en: *glad*, es: *<contento>* M. → *content\**; then *ii)* **Gender Accuracy** (GA), i.e., the proportion of words generated in the correct gender among the measurable ones, e.g., en:

---

[10] Available at: https://ict.fbk.eu/must-she/

[11] The test set also includes an audio portion aligned at the sentence level with its corresponding <transcript-translation> pair. However, as we are dealing with text-to-text translation, we do not rely on the audio portion of the MuST-SHE.

[12] As we will also discuss in §4.3, since automatic translation is an open-ended task, systems might generate a word that differs from the one annotated in MuST-SHE reference translation. As a result, all *out-of-coverage* words are necessarily left unevaluated by the automatic evaluation procedure.

*glad*, es: <*contento*> M. → *contento*. Hence, GA properly measures the tendency to (over)generalize masculine forms over feminine ones in automatic translation: scores below 50% can signal a strong bias, where the wrong gender form is generated by the systems more often than the correct one.

In our study, we rely on the above metrics to inspect gender translations and employ BLEU (Post, 2018) to measure overall translation quality so as to inspect how these two aspects relate to each other.

## 4. Results and Discussion

In our experiments, we compare the ability of GT, DL, and MMT (§3.1) to translate gender at several levels of granularity and how it relates to their overall translation quality. To do so, we access all systems via an API, feed them with MuST-SHE source English sentences as input,[13] and evaluate their automatic translations against MuST-SHE references following the evaluation method described in §3.2. Starting with a presentation of the systems' overall results and across gender forms (§4.1), we then proceed by inspecting their behaviour across different categories of phenomena (§4.3), and finally across word classes and POS (§4.4). Note that, as previously mentioned in §3.2, by relying on three target languages (es, fr, it) with similar typological features, we are able to verify under comparable conditions if the behaviour of the studied MT models is uniform. As such, we do not focus on the absolute results for each language pair in our discussions and evaluations, nor do we focus on the scores obtained by each system separately. Rather, we intend to observe if and to what extent biased behaviours are consistent across languages in their comparison across models.

### 4.1. Overall and Gender Results

Table 2 shows the results (i.e., percentage scores) achieved by the three MT models for overall translation quality in terms of BLEU scores, Term Coverage (All-Cov), and Gender Accuracy (All-Acc) on the complete MuST-SHE test set.

|       |                  | BLEU  | All-Cov | All-Acc |
|-------|------------------|-------|---------|---------|
| en-es | DeepL            | 44.66 | 77.8    | **76.3**|
|       | Google Translate | **46.22** | 79.7 | 73.4    |
|       | ModernMT         | 45.36 | **79.9**| 72.9    |
| en-fr | DeepL            | 40.27 | **72.9**| **73.4**|
|       | Google Translate | **40.94** | 72.2 | 72.5    |
|       | ModernMT         | 39.17 | 70.1    | 72      |
| en-it | DeepL            | **37.80** | **70.5** | **77.4** |
|       | Google Translate | 36.76 | 68.7    | 71.3    |
|       | ModernMT         | 36.87 | 69.2    | 71.8    |

Table 2: Overall quality scores (BLEU), Term Coverage, and Gender Accuracy on MuST-SHE.

Starting with BLEU scores[14] - despite variations across language pairs that are well-known in the MT community[15] - all systems unsurprisingly obtain state-of-the-art results indicating that the generated translations are of good overall quality. Whereas GT emerges as the best-performing system for en-es (46.22) and en-fr (40.94), it is DL that stands out for the en-it (37.8) language pair. Moving onto Coverage, instead, we can see a close relationship between the amount of MuST-SHE annotated words (regardless of their gender) generated by the systems and BLEU scores: systems with higher BLEU tend to have higher coverage. This emerges more predominantly across languages, where language pairs rank in the same order in terms of both BLEU and coverage (en-es > en-fr > en-it). For gender translation, however, we do not find the same pattern. In fact, accuracy scores rank differently across language pairs - en-it (77.4 DL) > en-es (76.3 DL) > en-fr (73.4 DL) - and show that DL achieved marginally better results than the other systems in our experiments in gender translation, despite not systematically being the top-performer according to the other metrics. Therefore, these first overall results already highlight how the generic translation quality of MT output neither directly relate to nor reflect the ability to translate gender properly.

---

[13] The automatic translations were collected November 4th, 2022.

[14] A score of 30-40 indicates an understandable/good translation, while 40-50 suggests a high-quality one. For more information, please refer to Vashee (2021b).

[15] Consistently across models, the highest results are achieved for en-es, followed by en-fr and en-it. These differences are also due to the varying amount of training language data available for each language pair.

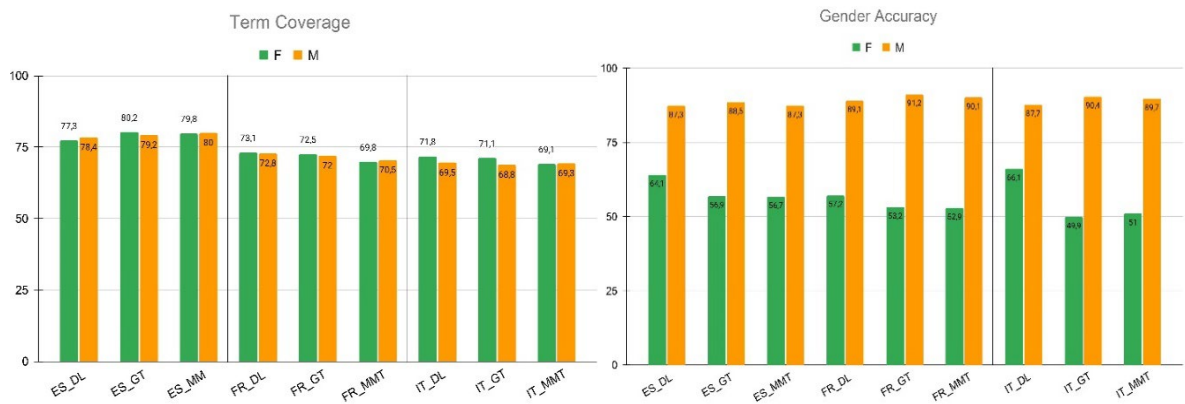

Figure 2: Feminine and masculine Coverage scores        Figure 3: Feminine and masculine Accuracy scores

To further investigate gender bias, however, we need to go beyond aggregated scores, and rather inspect a potential 'gender gap' between feminine and masculine translations. For this reason, we test and report both coverage (Figure 2) and accuracy (Figure 3) results distinguished by gender form. For coverage, Figure 2 shows how, within each language pair, words from the feminine and masculine sets are consistently generated at the same rate, meaning that we can measure gender accuracy on a comparable number of words. When it comes to assessing proper gender realization, however, Figure 3 attests to a substantial and systematic gender divide across gender forms (between ~23-40 accuracy points) for all languages and MT models, with masculine gender achieving around a very positive 90 points accuracy on average, and the feminine one getting to ~66 points at best. Noticeably, if we observe accuracy for feminine forms only, GT for en-it obtains the worst results (48.95), indicating that the system generates a wrong masculine form – instead of the expected feminine one - in more than half of the cases. Overall, while all systems are proved to be biased exhibiting a masculine skew, DL emerges as the more favoring system for feminine gender translation, whereas GT and MMT, which are basically on par, lag behind.

All in all, this first set of results and analyses reveals that all systems are affected by biased behaviors. Although concealed by BLEU and overall gender accuracy scores, results disaggregated by gender show a gender gap where feminine forms are always disfavoured. Among all systems, however, DL is the best one for (feminine) gender translations, thus attesting to the fact that higher/lower overall translation quality does not directly relate to better/worse gender capabilities.

## 4.2. Gender Categories

Our evaluation now becomes more fine-grained and accounts for the different categories of gender phenomena represented in MuST-SHE, namely, as described in §3.2, by distinguishing between i) gender-ambiguous input sentences from CAT1 (e.g., I am a friend), and ii) input sentences disambiguated via a contextual gender cue from CAT2 (e.g., He/she is a friend). On this basis, in Figures 4 and 5 we respectively show coverage and accuracy results across categories of phenomena, also disaggregated across feminine and masculine forms.

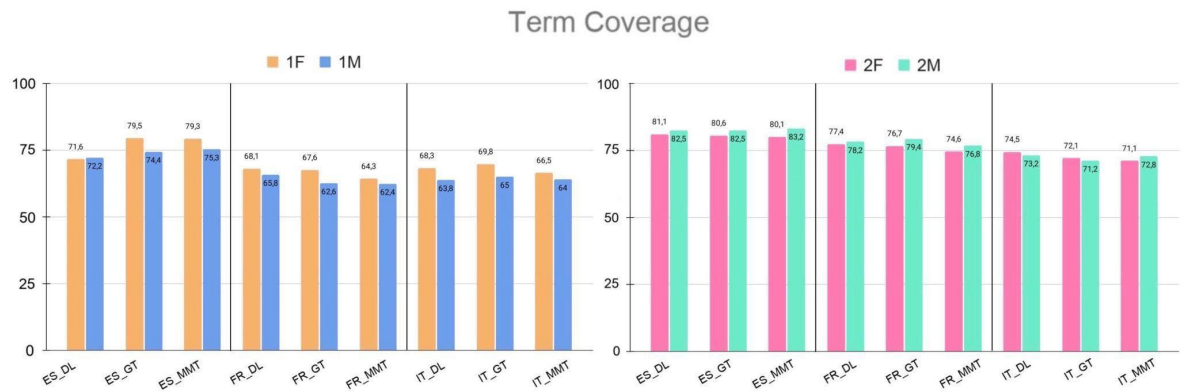

Figure 4: Coverage scores for all systems and language pairs, across Categories 1-2 and gender forms (F/M)

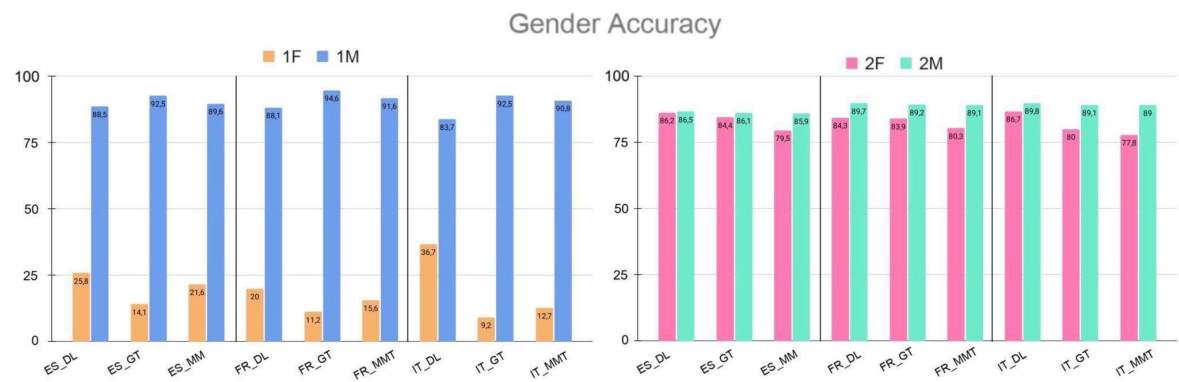

Figure 5: Accuracy scores for all systems and language pairs, across Categories 1-2 and gender forms (F/M)

Starting from Figure 4, we can see how – once again - coverage is generally on par across all systems, languages, and gender forms. The only attested difference, although it does not emerge as significant, is that words from CAT2 have slightly higher coverage compared to CAT1. Major differences across categories become instead evident in the accuracy results shown in Figure 5. By focusing on CAT2, we see that the gender gap is reduced (~10/12 points max). Masculine scores on the one hand are only slightly worse (compared to Figure 3), whereas feminine gender is consistently better on this set of sentences; for DL on en-es, results are even on par across genders. Overall, this attests to the systems' ability to rely on contextual cues to translate gender, even on natural sentences from MuST-SHE that can feature a high level of complexity, as shown in the following example (1).

(1) Sentence ID: it-0766

| | |
|---|---|
| **Source EN** | I was getting calls with questions like this from hundreds of women, **all concerned** that something was wrong, because they couldn't measure up. |
| **Ref IT** | Ho ricevuto telefonate con domande simili da centinaia di donne, **tutte preoccupate** che qualcosa stesse andando storto, perché non si sentivano all'altezza. [feminine] |
| **DL** | Ricevo telefonate con domande di questo tipo da centinaia di donne, **tutte preoccupate** che ci fosse qualcosa di sbagliato, perché non riuscivano ad essere all'altezza. [feminine] |
| **GT** | Ricevevo telefonate con domande come questa da centinaia di donne, **tutte preoccupate** che qualcosa non andava, perché non potevano essere all'altezza. [feminine] |
| **MMT** | Ricevevo chíamate con domande come questa da centinaia di donne, **tutte preoccupate** che qualcosa non andasse, perché non erano all'altezza. [feminine] |

Conversely, it is on phenomena from CAT1 that gender bias predominantly emerges: masculine translation can be correct at a rate as high as 94.6% (GT on en-fr) and feminine realization as low as 9.2% (GT on en-it). Even DL, which still systematically shows an edge for feminine translation compared to GT and MMT, does not manage to close the huge gap across gender forms, and still generates feminine terms at a rate below random chance (25.6 for en-es, 20 for en-fr, 36.7 for en-it). All in all, the gender divide demonstrated for this category of phenomena clearly signals that – when faced with ambiguous gender references – MT systems resort to a masculine translation almost by default.

This issue is exemplified by professional titles, included in CAT1-F. For this feminine category, those cases that were systematically generated in the wrong gender by all three systems concerned professions and/or roles that are stereotypically associated with men: *historian* (it-0142), *scientist* (it-0385; it-0386; it-0387; it-0459; it-0873; it-0963; it-0964; it-1049), *professor* (it-0512; it-0635), *researcher* (it-0635; it-0641), *expert* (it-0832), and *college graduate* (it-1069). Here follows one such case:

(2) Sentence ID: it-0385

| | |
|---|---|
| **Source EN** | As a **scientist**, I have access to high-tech equipment that we can put over the side of the research vessel, and it measures oxygen and many more things |
| **Ref IT** | Da **scienziata**, ho accesso ad attrezzature di alta tecnologia che si possono mettere sul bordo della nave da ricerca, per misurare il livello di ossigeno e molte altre cose. [feminine] |
| **DL** | In qualità di **scienziato**, … [masculine] |
| **GT** | Come **scienziato**, … [masculine] |
| **MMT** | Come **scienziato**, … [masculine] |

These scores emphasize the value of dedicated, fine-grained analyses, able to differentiate the behavior of each system across translation cases. With this in mind, we now move on to investigate if different word classes and POS are equally affected by gender bias.

## 4.3. Class and POS

Here, we evaluate the behavior of our MT models across content words from the open class (nouns, descriptive adjectives, verbs) and functional words from the closed class (limiting adjectives, pronouns, and articles/prepositional articles). Following Savoldi et al., (2022, p. 1810), since such classes of words differ substantially in terms of variability, frequency, and semantics, they can represent an interesting variable for gender translation.

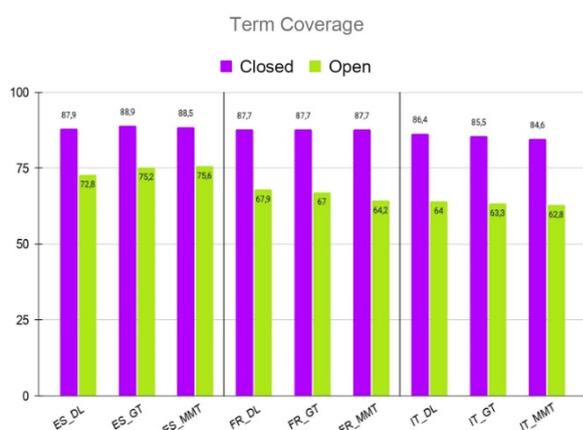

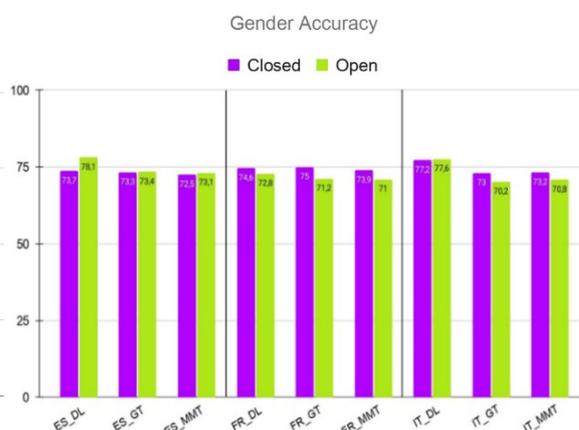

Figure 6: Coverage scores across for closed / open class

Figure 7: Accuracy scores across for closed / open class

Indeed, in Figure 6 we see that closed-class words systematically obtain a higher coverage compared to words from the open class. This is most likely because – compared to the higher complexity and sparse inflectional variants exhibited by words from the open class – closed-class POS have simple forms, less variability, and high frequency in the systems' training data. Thus, their correct generation is "easier" for translation models. Differently, in terms of gender translation accuracy shown in Figure 7, we find no noticeable contrasts: for all language pairs, all systems appear to perform quite similarly across classes, at least according to these gender-aggregated results. Hence, we take the next step to investigate gender translation for each POS, also distinguishing between feminine and masculine forms.

Figures 8 and 9 represent the coverage obtained by all systems on POS from the closed and open class, which are also disaggregated for feminine and masculine gender.

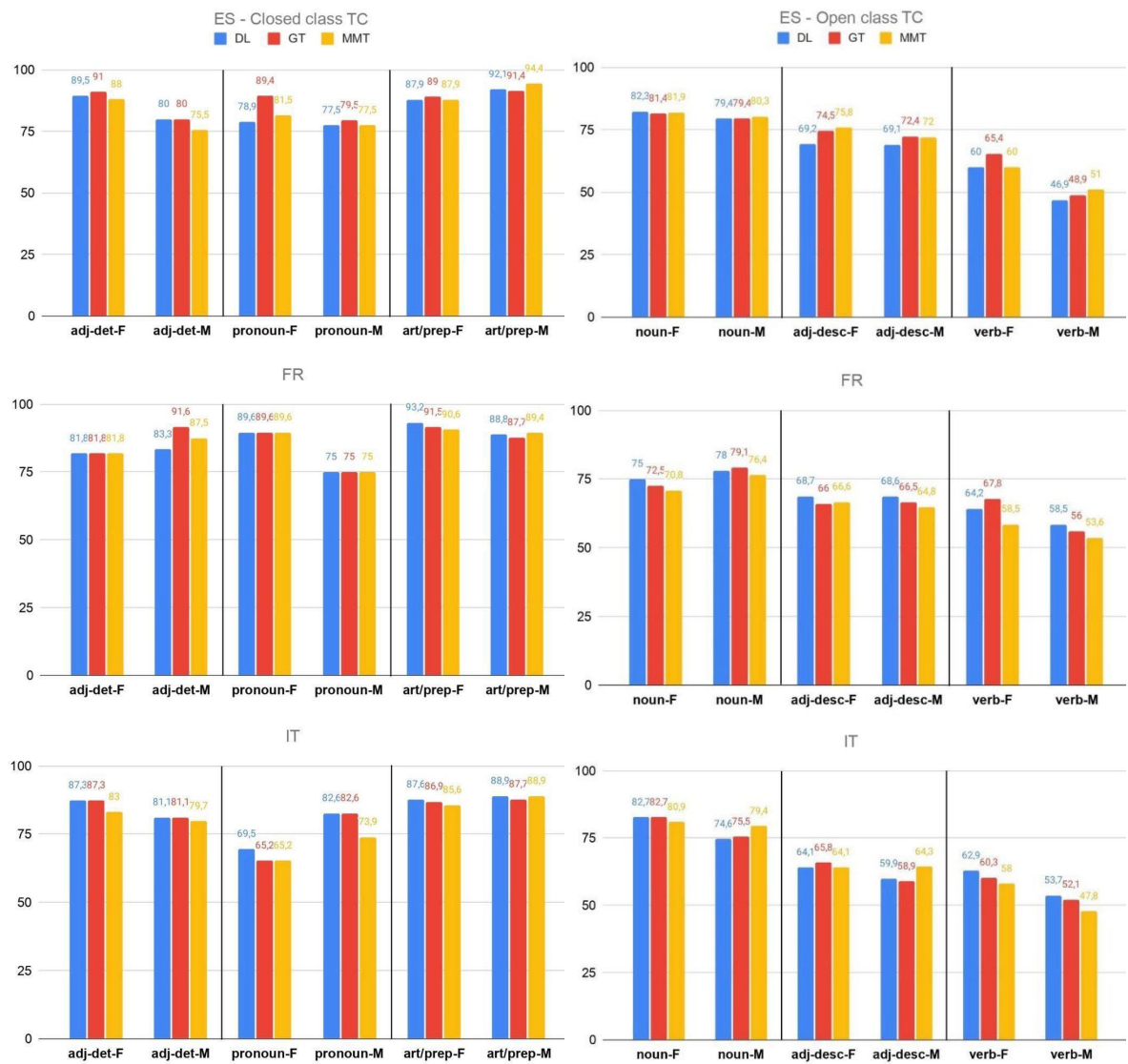

Figure 8: Coverage scores for POS from the closed class for feminine and masculine gender forms.

Figure 9: Coverage scores for POS from the open class for feminine and masculine gender forms.

For all languages and systems, coverage percentages are still overall quite similar across genders, with just small variations that do not appear systematic. Rather, we can also hereby confirm that – regardless of gender – POS from the open class (Figure 9) generally obtains lower coverage, as already attested in Figure 6. In particular, the most striking and negative result is obtained for VERB: almost 50% of MuST-SHE annotated verbs are not generated in the MT output (Figure 9). This is consistent across languages and systems, and more evident for the masculine set. For these cases, systems generated an alternative, but equally valid verb, i.e., the same verb with a form change, such as in tense/voice, or another synonymic verb.[16] In line with the findings of Savoldi et al., (2022, p. 1814), such behavior seems quite frequent, and the following is an example of such cases.[17]

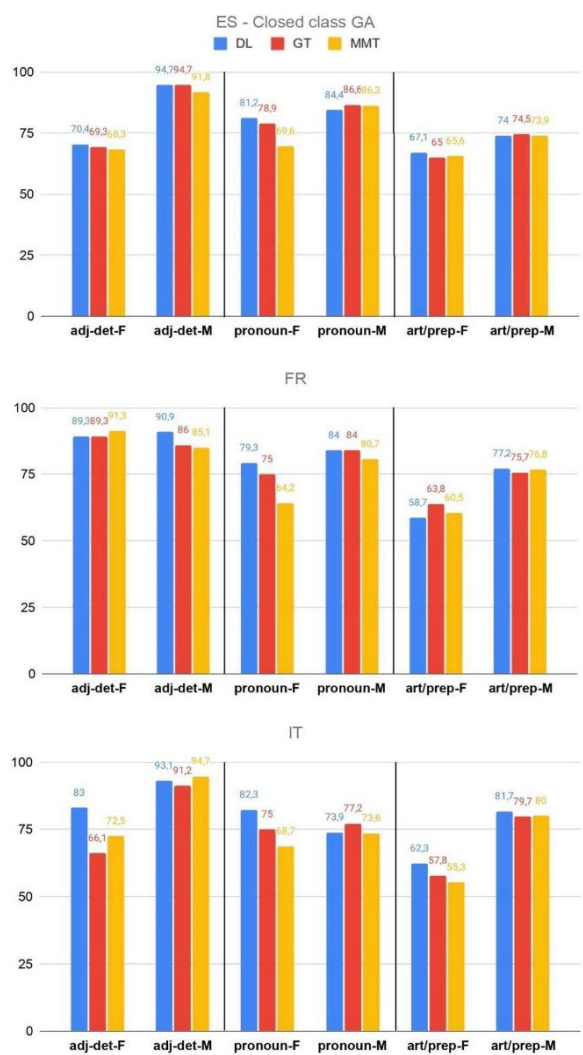

Figure 10: Accuracy scores for Closed POS across feminine and masculine forms.

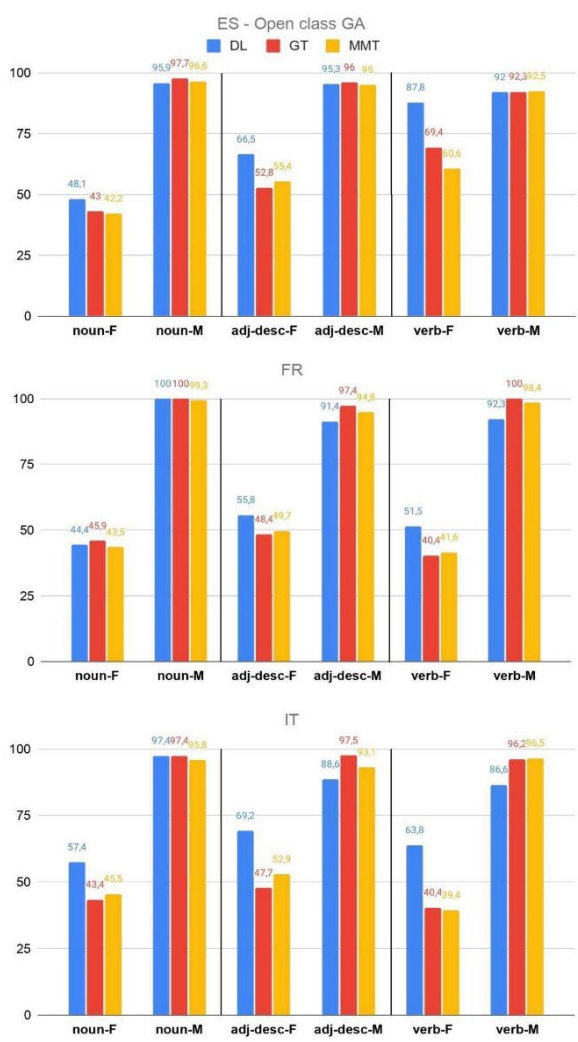

Figure 11: Accuracy scores for Open POS across feminine and masculine forms.

(3) Sentence ID: it-0134

| | |
|---|---|
| **Source EN** | So I'm going to show you exactly what I **brought**. |
| **Ref IT** | Voglio mostrarvi esattamente cosa *mi sono* **portata**. [feminine] |
| **DL** | Quindi vi mostrerò esattamente cosa *ho portato*. |
| **GT** | Quindi ti mostrerò esattamente cosa *ho portato*. |
| **MMT** | Quindi ti mostrerò esattamente quello che ti *ho portato*. |

---

[16] This insight is derived from a manual overview of ~100 output Italian verbs generated by the three systems.

[17] This insight is derived from a manual overview of ~100 output Italian verbs for the three systems.

In the reference, we find a reflexive form for the verb *portare* (to bring), which has a past participle (*portata*) that must agree in gender (feminine) and number (singular) with the subject (the speaker). In the output, however, the systems generated the active form of the verb, which requires number agreement but is not gender marked. The translation proposed by the systems is equally correct, but the automatic evaluation script does not manage to match the word. This could therefore explain why the automatic evaluation reports such a low percentage of Term Coverage for the VERB category.

Moving on to accuracy scores, we show closed (Figure 10) and open (Figure 11) class POS results across gender forms. By comparing such Figures, we unveil that the most severe bias is attested for open POS, which presents wider gender gaps across masculine and feminine forms (in most cases by even more than 50 points; Figure 11) with respect to the closed ones (max ~20 points, Figure 10). Additionally, nouns arguably appear as the most biased POS, with the biggest divide across gender forms. Indeed, feminine nouns (NOUN-F) largely remain below 50% accuracy, with DL for en-it representing the only exception (57.4% − Figure 11). Differently, the gender realization of masculine nouns (NOUN-M) is basically perfect, as they are correctly translated in above 95% of the cases. Indeed, there are only very few instances (4-5 words for en-es/it) where systems wrongly generate a feminine translation of a NOUN-M. These correspond to biased output which contains stereotypes or controversial gendered associations, as shown in the following example (4) with the world 'liar' translated as *bugiarda* (feminine) rather than *bugiardo* (masculine). In this case, DeepL was the only system to generate not only a correct output, but also one of higher quality and fluency compared to the other two.

(4) Sentence ID: it-0017

| | |
|---|---|
| **Source EN** | He said to me, "I had to give this prognosis to your parents that you would never walk, and you would never have the kind of mobility that other kids have or any kind of life of independence, and you've been making **liar** out of me ever since." |
| **Ref IT** | Mi ha detto: "ho dovuto dare questa prognosi ai tuoi genitori che tu non avresti mai camminato, e che non avresti mai avuto il tipo di mobilità che hanno gli altri bambini o nessun tipo di vita indipendente, e voi mi avete dato del *bugiardo* da allora" [masculine] |
| **DL** | Mi disse: "Ho dovuto dare ai tuoi genitori la prognosi che non avresti mai camminato e non avresti mai avuto il tipo di mobilità che hanno gli altri bambini o una vita indipendente, e da allora mi hai fatto passare per un **bugiard<u>o</u>**". [masculine] |
| **GT** | Mi ha detto: "Ho dovuto dare questa prognosi ai tuoi genitori che non avresti mai camminato, e non avresti mai avuto il tipo di mobilità che hanno gli altri bambini o qualsiasi tipo di vita di indipendenza, e stai facendo la **bugiard<u>a</u>** di me da allora" [feminine] |
| **MMT** | Mi disse: "Ho dovuto dare questa prognosi ai tuoi genitori che non avresti mai camminato, e non avresti mai avuto il tipo di mobilità che hanno gli altri bambini o qualsiasi tipo di vita di indipendenza, e da allora mi stai facendo diventare una **bugiard<u>a</u>**" [feminine] |

Overall, our results are in line with previous analyses on commercial (Monti, 2017; Prates et al., 2018; Rescigno et al., 2020) as well as non-commercial MT systems (Savoldi et al., 20220), showing how professions represent an arduous lexical category for gender translation, the one most affected by gender bias.

## 4.4. Final Remarks

Considering the above, our experiments and findings shed a spotlight on the persistent issue of gender bias within MT systems. Such bias not only hampers system performance and reliability but also carries profound societal and ethical implications. In fact, biased MT models pose a tangible risk of perpetuating harmful gender stereotypes – particularly in the context of professions, as our findings have demonstrated. Considering that DL, GT, and MMT are utilized by millions of people worldwide, their biased translations could inadvertently establish a norm for users seeking translations, even if implicitly. Indeed, MT systems that disproportionately favor male-centric language and masculine occupational terms not only reflect the existing societal asymmetries; rather they also have the potential to normalize the use of male-centric titles while linguistically excluding the representation of women in high-prestige careers. Furthermore, the limited capability of these systems to provide feminine translations can lead to immediate disadvantages for women relying on MT, for instance, as they might need to correct multiple – wrongly translated – feminine references when referring to themselves. As highlighted in the previous section, this issue is especially pronounced for ambiguous sentences lacking gender cues (i.e., CAT1), where DL, GT, and MMT consistently default to masculine forms. Instead, we have found that unambiguous input sentences (i.e., CAT2) are provided with the correct gender translations at a consistently higher rate.

As such, it is important to note that these systems do not fail uniformly across all linguistic phenomena, and our findings offer valuable insights for informing and alerting users of MT systems. As massively deployed tools, we believe it is indeed crucial to highlight their shortcomings as well as the risk they might bring to users, such as by reducing feminine visibility in language and potentially impeding effective communication. Additionally, and on a more positive note, our study can serve as a catalyst for proactive measures. For instance, in the case of ambiguous inputs like CAT1, the lack of contextual cues for gender simply hinders the possibility for automatic systems of providing a single, correct gender translation. Therefore, our findings corroborate and encourage existing research focusing on the development of technical solutions that offer dual gender outputs (e.g., Alhafni et al., 2023; Sánchez et al., 2023), which can empower users to select the correct gender form according to their preferences and needs. This approach can not only enhance the usability of MT systems but also promote inclusivity and gender sensitivity in translation technology.

## 5. Conclusion

To conclude, in this paper we have shed light on the implications of gender translation and bias for three commercial MT systems: DeepL, Google Translate, and ModernMT. By means of focused evaluations on three language pairs (en-es/it/fr) and at different levels of granularity, we have shown how these systems tend to be biased and under-represent feminine forms, particularly when confronted with ambiguous translations and occupational nouns.

Also, we consistently confirmed that gender translation ability does not necessarily correlate with overall translation quality. In fact, despite not always being the most competitive model in terms of generic translation quality, DeepL emerges as the system that better handles feminine gender translation.

Overall, this dedicated gender analysis highlights the importance of taking into account gender influence when assessing the quality of MT and takes stock of how bias issues still affect popular, commercial MT tools available online. Finally, we conclude by emphasizing how our findings could inform MT users' literacy on the design of technical countermeasures.

## Acknowledgement

Silvia Alma Piazzolla's work was carried out during an internship at the Fondazione Bruno Kessler. Furthermore, we acknowledge the support of the PNRR project FAIR - Future AI Research (PE00000013), under the NRRP MUR program funded by the NextGenerationEU.